\documentclass{llncs}
\usepackage{llncsdoc}
\usepackage{latexsym}
\usepackage{amsmath} 

\usepackage[pdftex]{graphicx} 
\usepackage{subfigure}
\usepackage{epstopdf}
\usepackage{pinyin}
\usepackage{CJK} 

\usepackage{multirow} 
\usepackage{array}

\title{Word, Subword or Character? An Empirical Study of Granularity in Chinese-English NMT}

\author{Yining Wang$^{\dagger}$, Long Zhou$^{\dagger}$, Jiajun Zhang$^{\dagger}$, Chengqing Zong$^{\dagger\ddagger}$ \\}
\institute{
  $^\dagger$University of Chinese Academy of Sciences \\
  National Laboratory of Pattern Recognition, CASIA  \\
  $^\ddagger$CAS Center for Excellence in Brain Science and Intelligence Technology \\
  {\tt \{yining.wang,long.zhou,jjzhang,cqzong\}@nlpr.ia.ac.cn} \\}

\begin{document}

\begin{CJK*}{UTF8}{gbsn}

\maketitle

\begin{abstract}
  Neural machine translation (NMT), a new approach to machine translation, has been proved to outperform conventional statistical machine translation (SMT) across a variety of language pairs.
  Translation is an open-vocabulary problem, but most existing NMT systems operate with a fixed vocabulary, which causes the incapability of translating rare words. This problem can be alleviated by using different translation granularities, such as character, subword and hybrid word-character. Translation involving Chinese is one of the most difficult tasks in machine translation, however, to the best of our knowledge, there has not been any other work exploring which translation granularity is most suitable for Chinese in NMT. In this paper, we conduct an extensive comparison using Chinese-English NMT as a case study. Furthermore, we discuss the advantages and disadvantages of various translation granularities in detail. Our experiments show that subword model performs best for Chinese-to-English translation with the vocabulary which is not so big while hybrid word-character model is most suitable for English-to-Chinese translation. Moreover, experiments of different granularities show that Hybrid\_BPE method can achieve best result on Chinese-to-English translation task.

\end{abstract}

\section{Introduction}

Neural machine translation (NMT) proposed by Kalchbrenner and Blunsom~\cite{Kalchbrenner:2013} and Sutskever et al.~\cite{Sutskever:2014} has achieved significant progress in recent years.
Unlike traditional statistical machine translation(SMT) \cite{Koehn:2003,Chiang:2005,zhai2012tree} which contains multiple separately tuned components, NMT builds an end-to-end framework to model the entire translation process. For several language pairs, NMT has already achieved better translation performance than SMT \cite{Wu:2016,Junczys-Dowmunt:2016}.

Conventional NMT system limits the vocabulary to a modest-sized vocabulary in both sides and words out of vocabulary are replaced by a special \textbf{UNK} symbol. However, the process of training and decoding is often conducted on an open vocabulary, in which an obvious problem is that NMT model is incapable of translating rare words. In particular, if a source word is outside the source vocabulary or its translation is outside the target vocabulary, the model is unable to generate proper translation for this word during decoding. Both Sutskever et al.~\cite{Sutskever:2014} and Bahdanau et al.~\cite{Bahdanau:2015} have observed that sentences with many out-of-vocabulary words tend to be translated much more poorly than sentences mainly containing frequent words.

To address this problem, many researchers propose a broad category of approaches by employing different translation granularities. Most of these are below the word level, e.g.\ characters \cite{Chung2016A}, hybrid word-characters \cite{Luong2016Achieving,Wu:2016}, and more intelligent subwords \cite{Sennrich:2016A,Wu:2016}.
Besides, pioneering studies~\cite{Wu:2016,Junczys-Dowmunt:2016} demonstrate that translation tasks involving Chinese are some of the most difficult problems in NMT systems.
However, there is no study that shows which translation granularity is suitable for Chinese-to-English and English-to-Chinese translation tasks.

In this work, we make an empirical comparison of different translation granularities for bidirectional English-Chinese translation tasks. In addition, we analyze the impact of these strategies on the translation results in detail. We demonstrate that Chinese-to-English NMT of 15k and 30k vocabulary size can acquire best results using subword model and with 60k vocabulary size hybrid word-character model obtains the highest performance, while hybrid word-character model is most suitable for English-to-Chinese translation. Our experiment shows that all subword methods are not bounded by the vocabulary size. Furthermore, we carry out the experiments that employ different translation granularities of source side and target side. The translation result shows that when the source granularity is hybrid word-character level and the target sentences are split into subword level by BPE method, it can achieve the best translation performance for Chinese-to-English translation task. As for English-to-Chinese translation task, Hybrid word-character model is most suitable. To the best of our knowledge, this is the first work on an empirical comparison of various translation granularities for bidirectional Chinese-English translations.

\begin{figure*}[t]
    \centering
    \includegraphics[width=11cm]{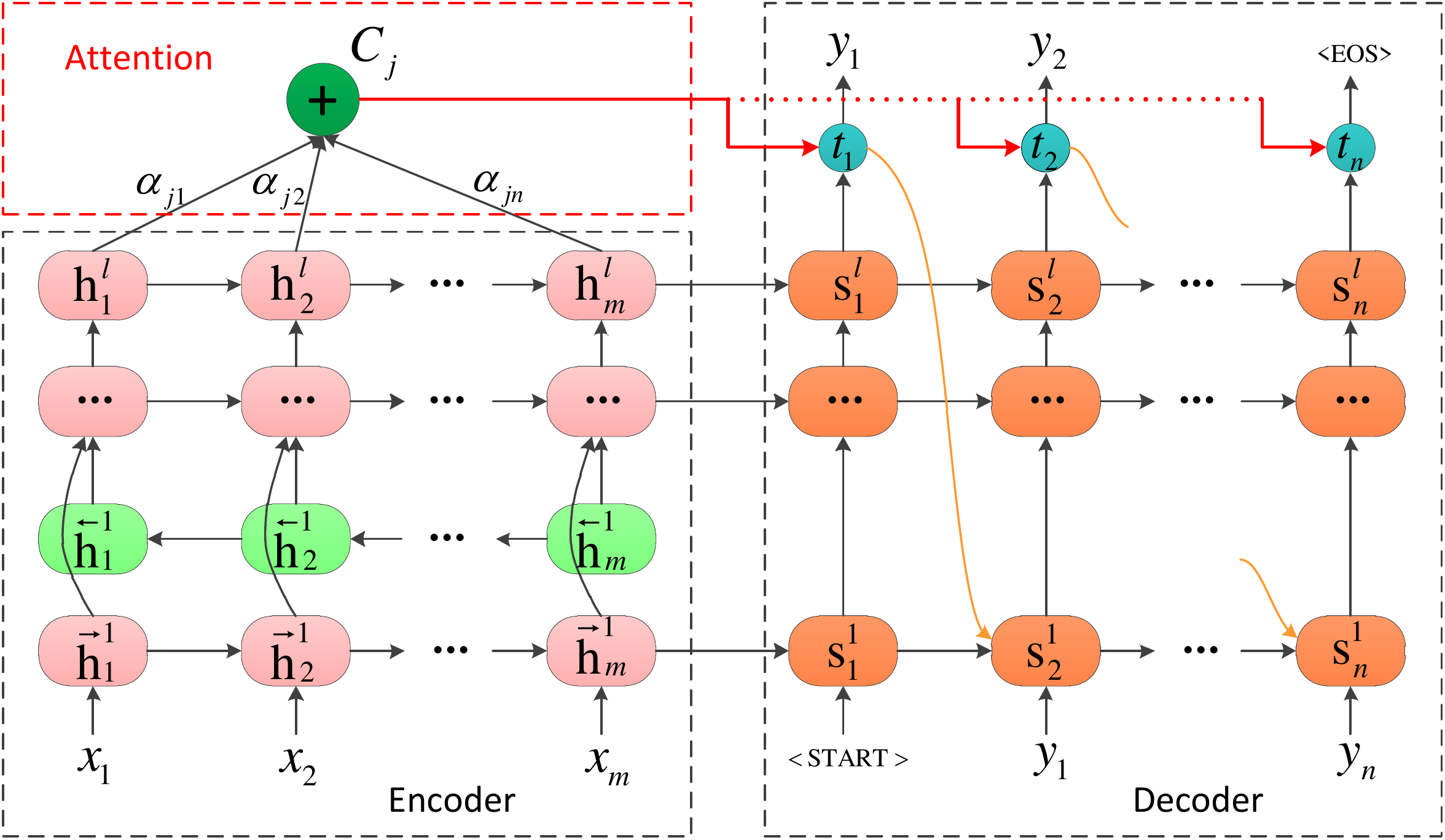}
    \caption{The architecture of neural machine translation model.}\label{fig:1}
\end{figure*}

\section{Neural Machine Translation} \label{nmt}

Our models are based on an encoder-decoder architecture with attention mechanism proposed by Luong et al.~\cite{Luong:2015A}, which utilizes stacked LSTM layers for both encoder and decoder as illustrated in Figure \ref{fig:1}. In this section, we make a review of NMT framework.

First, the NMT encodes the source sentence $X=(x_1,x_2,...,x_m)$ into a sequence of context vector representation $C=(h_1,h_2,...,h_m)$. Then, the NMT decodes from the context vector representation $C$ and generates target translation $Y=(y_1,y_2,...,y_n)$ one word each time by maximizing the probability of $p(y_j|y_{<j},C)$.
Next, We review the encoder and decoder frameworks briefly.

{\bf Encoder:} The context vector representation $C=(h_1^l,h_2^l,...,h_m^l)$ is generated by the encoder using $l$ stacked LSTM layers.
Bi-directional connections are used for the bottom encoder layer, and $h_i^1$ is a concatenation vector as shown in Eq. (1): 
\begin{equation}
\label{equ:Eq1}
    h_i^1 = \begin{bmatrix} \overrightarrow{h}_i^1 \\ \overleftarrow{h}_i^1 \end{bmatrix} =
    \begin{bmatrix} LSTM(\overrightarrow{h}_{i-1}^1, x_i) \\ LSTM(\overleftarrow{h}_{i-1}^1,x_i) \end{bmatrix}
\end{equation}

All other encoder layers are unidirectional, and $h_i^k$ is calculated as follows:
\begin{equation}
    h_i^k = LSTM(h_{i-1}^k, h_i^{k-1})
\end{equation}

{\bf Decoder:} The conditional probability $p(y_j|y_{<j},C)$ is formulated as
\begin{equation} \label{condi}
    p(y_j|Y_{<j},C) = p(y_j|Y_{<j},c_j) = softmax(W_st_j)
\end{equation}

Specifically, we employ a simple concatenation layer to produce an attentional hidden state $t_j$:
\begin{equation} \label{concat}
    t_j = tanh(W_c[s_j^l;c_j]+b) = tanh(W_c^1s_j^l+W_c^2c_j+b)
\end{equation}
where $s_j^l$ denotes the target hidden state at the top layer of a stacking LSTM.
The attention model calculates $c_j$ as the weighted sum of the source-side context vector representation, just as illustrated in the upper left corner of Figure \ref{fig:1}.
\begin{equation}
    c_j = \sum_{i=1}^{m}ATT(s_j^l,h_i^l) \cdot h_i^l =  \sum_{i=1}^{m}\alpha_{ji}h_i^l
\end{equation}
where $\alpha_{ji}$ is a normalized item calculated as follows:

\begin{equation}
    \alpha_{ji} = \frac{exp(h_i^{l^T} \cdot s_j^l)}{\sum_{i^{'}}exp(h_{i^{'}}^{l^T} \cdot s_j^l)}
\end{equation}

$s_j^k$ is computed by using the following formula:
\begin{equation}
    s_j^k = LSTM(s_{j-1}^k, s_j^{k-1})
\end{equation}

If $k = 1$, $s_j^1$ will be calculated by combining $t_{j-1}$ as feed input~\cite{Luong:2015A}:
\begin{equation}
    s_j^1 = LSTM(s_{j-1}^1, y_{j-1}, t_{j-1})
\end{equation}

Given the bilingual training data $D = \{(X^{(z)},Y^{(z)})\}_{z=1}^Z$, all parameters of the attention-based NMT are optimized to maximize the following conditional log-likelihood:
\begin{equation}
    L(\theta) = \frac{1}{Z} \sum_{z=1}^{Z} \sum_{j=1}^n log p(y_j^{(z)} | y_{<j}^{(z)}, X^{(z)}, \theta)
\end{equation}

\section{Description of Different Translation Granularities}

We revisit how the source and target sentences ($X$ and $Y$) are represented in NMT. For the source side of any given training corpus, we scan through the whole corpus to build a vocabulary $V_x$ of unique tokens. A source sentence $X=(x_1,x_2,...,x_m)$ is then built as a sequence of the integer indices. The target sentence is similarly transformed into a target sequence of integer indices.

The property of NMT allows us great freedom in the choice of token units, and we can segment sentences in different ways. In this section, we will elaborate on four proposed approaches about the choice of translation granularities.

\begin{figure}[!t]
	\centering
	\includegraphics[width=0.9 \columnwidth]{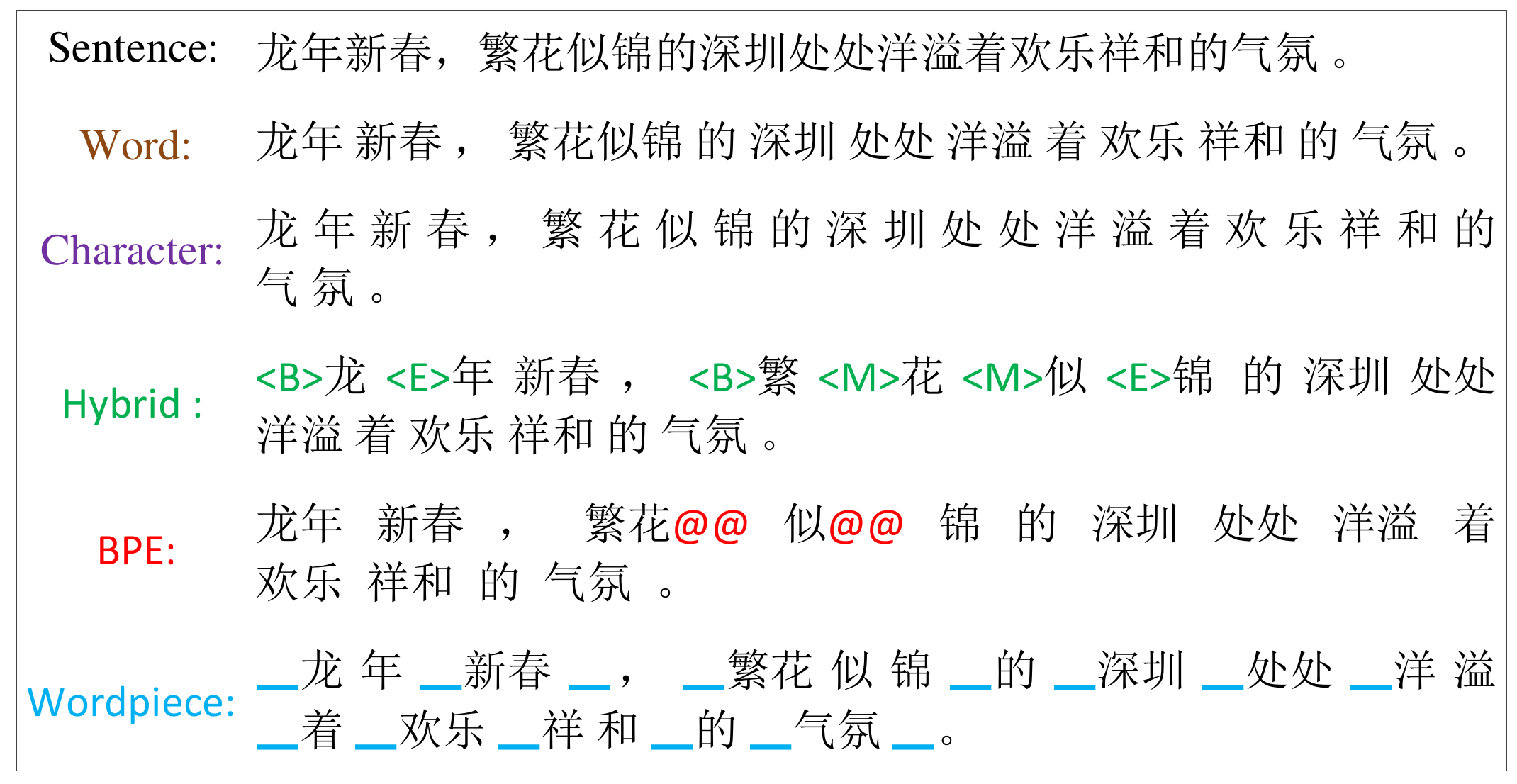}
	\caption{An example of different translation granularities}
	\label{frame}
\end{figure} 

\subsection{Character Level}
This translation granularity is easy to implement. For this granularity, what we have to do is split the sentence into a sequence of characters. However, the character-level modeling on the English side is more challenging, as the network has to be able to deal with long and coherent sequence of characters. In this case, the number of characters is often 300$\sim$1000 symbols long, where the size of the state space grows exponentially. Therefore, this is a great challenge for us to handle. 

Besides, the alphabet of English is only consist of 26 letters, in which the vocabulary of English side is too small. Considering these facts, we only separate the Chinese side sentences into characters rather than both sides. Figure~\ref{frame} shows an example of this translation granularity for character level. 

\subsection{Hybrid Word-Characters Level}
In regular word-based NMT, for all words outside the source vocabulary, one feeds the universal embedding representing \textbf{UNK} as input to the encoder. This is problematic because it discards valuable information about the source word. To address that, hybrid word-character approach will be adopted. In this part, we will introduce this granularity in detail.

Unlike in the conventional word model where out-of-vocabulary words are collapsed into a single \textbf{UNK} symbol, we convert these words into the sequence of constituent characters. Special prefixes are prepended to the characters. The purpose of the prefixes is to show the location of the characters in a word, and to distinguish them from normal in-vocabulary characters. There are three prefixes: \textbf{$\langle$B$\rangle$}, \textbf{$\langle$M$\rangle$}, and \textbf{$\langle$E$\rangle$}, indicating beginning of the word, middle of the word and end of the word, respectively. During decoding, the output may also contain sequences of special tokens. With the prefixes, it is trivial to reverse the tokenization to the original words as part of a post-processing step. Using this approach, in Figure~\ref{frame}, we can see the word ``龙年'' is segmented into ``$\langle$B$\rangle$龙 ~ $\langle$E$\rangle$年'', and the word ``繁花似锦'' is segmented into ``$\langle$B$\rangle$繁 ~$\langle$M$\rangle$花 ~ $\langle$M$\rangle$似~ $\langle$E$\rangle$锦''. 

\subsection{Subword Level}
Considering languages with productive word formation processes such as agglutination and compounding, translation models require mechanisms that segment the sentence below the word level (In this paper, we call this level of symbols as subword units). In this part, we will introduce the two different methods of translation granularity on subword level.
\subsubsection{BPE Method}

Byte pair encoding (BPE) \cite{Gage1994A} is a compression algorithm. This simple data compression technique iteratively replaces the most frequent pair of bytes in a sequence with a single, unused byte. This compression method is first introduced into translation granularity by Sennrich et al.~\cite{Sennrich:2016A}. In this approach, instead of merging frequent pairs of bytes, characters or character sequences will be merged.

A detailed introduction of algorithm in learning BPE operations is showed in Sennrich et al.~\cite{Sennrich:2016A}. During decoding time, each word first split into sequences of characters, then learned operation will be applied to merge the characters into larger, known symbols. For BPE method, a special symbol is also needed to indicate the merging position. In Figure~\ref{frame}, the word ``繁花似锦'' is segmented into three subword units, and the first three units are appended a special suffix ``@@''. In decoding step, the translation results contain the special tokens as well. With these suffixes, we can recover the output easily. 
\subsubsection{WPM Method}

The wordpiece model (WPM) implementation is initially developed to solve a Japanese/Korean segmentation problem for the speech recognition system \cite{Schuster2012Japanese}. This approach is completely data-driven and guaranteed to generate a deterministic segmentation for any possible sequence of characters, which is similar to the above method.

The wordpiece model is generated using a data-driven approach to maximize the language-model likelihood of the training data, given an evolving word definition. The training method of WPM is described in more detail in Schuster and Nakajima \cite{Schuster2012Japanese}. As shown in Figure~\ref{frame}, a special symbol is only prepended at the beginning of the words. In this case, the words ``龙年'', ``繁花似锦'', ``洋溢'' and ``祥和'' are split into subwords, and the rest words remain the same except for a special prefix ``\_''.

\section{Experiments}

\subsection{Dataset}

We perform all these translation granularities on the NIST bidirectional Chinese-English translation tasks. The evaluation metric is BLEU~\cite{Papineni:2002} as calculated by the {\small\tt multi-bleu.perl} script.

Our training data consists of 2.09M sentence pairs extracted from LDC corpus{\footnote[1]{The corpora include LDC2000T50, LDC2002T01, LDC2002E18, LDC2003E07, LDC2003E14, LDC2003T17 and LDC2004T07.}}. Table 1 shows the detailed statistics of our training data. To test different approaches on Chinese-to-English translation task, we use NIST 2003(MT03) dataset as the validation set, and NIST 2004(MT04), NIST 2005(MT05), NIST 2006(MT06) datasets as our test sets. For English-to-Chinese translation task, we also use NIST 2003(MT03) dataset as the validation set, and NIST 2008(MT08) will be used as test set.

\begin{table}[htbp]
\centering
\caption{\label{tab:test} The characteristics of our training dataset on the LDC corpus.}
\begin{tabular}{|p{2.2cm}<{\centering}|p{1.5cm}<{\centering}|p{1.5cm}<{\centering}|p{1.5cm}<{\centering}|}

\hline
Corpora & & Chinese & English\\
 \hline
 \hline
 \multirow{3}{*}{LDC corpora}& \#Sent. & \multicolumn{2}{c|}{2.09M} \\
 \cline{2-4}
 &\#Word&43.14M & 47.73M \\
  \cline{2-4}
 &Vocab & 0.39M & 0.23M \\
  \hline
 \end{tabular}

\end{table}

\subsection{Training Details}
We build the described models modified from the Zoph\_RNN{\footnote[2]{https://github.com/isi-nlp/Zoph\_RNN}} toolkit which is written in C++/CUDA and provides efficient training across multiple GPUs. Our training procedure and hyper parameter choices are similar to those used by Luong et al.~\cite{Luong:2015A}. In the NMT architecture as illustrated in Figure \ref{fig:1}, the encoder has three stacked LSTM layers including a bidirectional layer, followed by a global attention layer, and the decoder contains two stacked LSTM layers followed by the softmax layer.

The word embedding dimension and the size of hidden layers are all set to 1000. We limit the maximum length in training corpus to 120. Parameter optimization is performed using both stochastic gradient descent(SGD) method and Adam method \cite{Kingma2014Adam}. For the first three epoches, We train using the Adam optimizer and a fixed learning rate of 0.001 without decay. For the remaining six epoches, we train using SGD, and we set learning rate to 0.1 at the beginning and halve the threshold while the perplexity go up on the development set. We set minibatch size to 128. Dropout was also applied on each layer to avoid over-fitting, and the dropout rate is set to 0.2. At test time, we employ beam search with beam size b = 12.
\subsection{Data Segmentation}
For Chinese word segmentation, we use our in-house segmentation tools. For English corpus, the training data is tokenized with the Moses tokenizer. We carry out Chinese-to-English translation experiment on 30k vocabulary and 15k vocabulary for both sides respectively, and we also conduct English-to-Chinese translation experiment on 30k vocabulary size. The word level translation granularity is set to our baseline method. 

For character level, we only segment the Chinese sentences into characters and the English sentences remain the same. For hybrid word-characters level, we segment training corpus for both sides. We rank the word frequency from greatest to least in training corpus, and in order to prevent the pollution from the very rare word, we have to set a segmentation point relatively higher. For 30k vocabulary, the word frequency below 64 is segmented into characters on Chinese side, and the segmentation point is set to 22 on English side. For 15k vocabulary, we set the segmentation point to 350 and 96 on Chinese side and English side respectively. For 60k vocabulary, the frequency of Chinese words below 14 and that of English words below 6 are split into characters.

For subword level, two different approaches are used. In BPE method{\footnote[3]{https://github.com/rsennrich/subword-nmt}}, the number of merge operations is set to 30000 on 30k vocabulary size, 15000 on 15k vocabulary size and 60000 on 60k vocabulary size. For Chinese sentences, we segment the training corpus using our in-house segmentation tools first, and then we can apply the BPE method same as English sentences. Considering the essence of WPM method{\footnote[4]{https://github.com/google/sentencepiece}}, we do not have to segment words for Chinese and tokenize sentences for English. That is to say, we can train the WPM without pre-processing step. Hence, for WPM method, we conduct our experiments both on the sentences trained on the raw corpus and the sentences trained on the segmented corpus.

\subsection{Results on Chinese-to-English Translation}
\subsubsection{30k Vocabulary Size}
We list the BLEU scores of different translation granularities on 30k vocabulary in Table \ref{c2e}.

\begin{table}
\centering
\caption{Translation results (BLEU score) of 30k vocabulary for Chinese-to-English translation.} \label{c2e}
\begin{tabular}{p{4.3cm}|p{1.4cm}<{\centering}p{1.4cm}<{\centering}p{1.4cm}<{\centering}p{1.4cm}<{\centering}|p{1.4cm}<{\centering}}
  \hline
  Segmentation (30k)              &    MT03(dev) &  MT04   &  MT05   &   MT06   & Ave \\
  \hline
  \hline
  Word level               &    41.48   &   43.67   &   41.37  & 41.92  & 42.11 \\
  Character level                &    42.72   &   44.12   &   41.29  & 41.83  & 42.49 \\
  Hybrid word-characters level   & 43.24   &      45.18   &   {\bf 42.96}   &   42.89 &  43.57  \\
  BPE method   &    43.78   &   {\bf 45.47}   &    42.37  & {\bf 43.37} & {\bf 43.75} \\
  WPM method (raw)      &  41.96 &   43.38 & 40.84 &  40.98  & 41.79\\
  WPM method    & {\bf 44.12} &   44.96 &  42.34 &  42.18  & 43.40 \\
  
  \hline
\end{tabular}
\end{table}

Row 1 is translation result of the state-of-the-art NMT system with word level. For the character level granularity (Row 2), the translation quality is higher than the word level by only 0.38 BLEU points. The last three lines in Table \ref{c2e} are subword level translation granularity, which contains BPE method and WPM method. BPE method (Row 4) achieves the best translation performance, which gets an improvement of 1.64 BLEU points over the word level. As for the WPM method (Row 6), the gap between this method and BPE method is narrow. Moreover, hybrid word-character level model (Row 3) outperforms the word level by 1.46 BLEU points, and translation quality of this method is very close to the BPE method. Experiments show that hybrid word-character level granularity and BPE method of subword level granularity are our choices for translation granularity on Chinese-to-English translation task.

\subsubsection{Comparison in Sentences of Different Lengths}
We execute different translation granularities on the training corpus. To make a comparison, We randomly choose 10000 sentences. Table~\ref{length} show the average sentence length of different methods on all granularities.

\begin{table}
\centering
\caption{Sentence length of different translation granularities.} \label{length}
\begin{tabular}{p{2.7cm}|p{1.4cm}<{\centering}p{1.4cm}<{\centering}p{1.4cm}<{\centering}p{1.4cm}<{\centering}p{1.4cm}<{\centering}p{1.58cm}<{\centering}}
  \hline
  Language              &    word &  character   &  Hybrid   &  BPE  & WPM   & WPM(raw) \\
  \hline
  \hline
  Source(Chinese)   &    20.60        &  33.84     &  22.07    &  21.56  & 22.13  & 18.17 \\
  Target(English)   &  22.85     &   22.85   &  25.00   & 23.52 & 24.43 & 23.85\\
  \hline
\end{tabular}
\end{table}

\begin{figure}
    \centering
    \includegraphics[width=10cm]{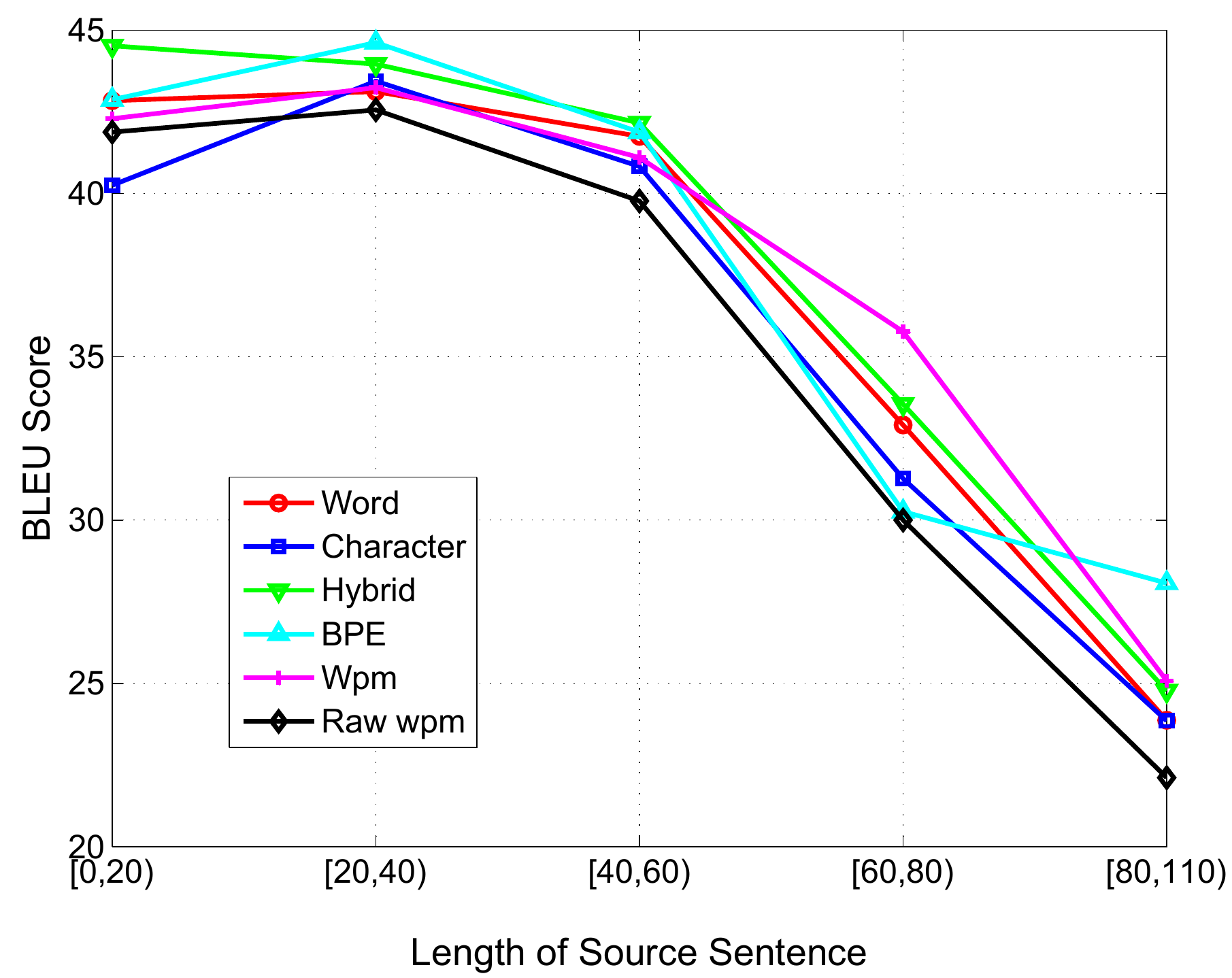}
    \caption{{\bf Length Analysis} - translation qualities(BLEU score) of different lengths.}\label{fig:3}
\end{figure}

A well-known flaw of NMT model is the inability to properly translate long sentences. However, most of translation granularities will go below the word level. Therefore, as shown in Table~\ref{length}, we can get longer sentences than the word level. We wonder what the translation performance of different lengths are on all translation granularities. We follow Bahdanau et al.~\cite{Bahdanau:2015} to group sentences of similar lengths together and compute a BLEU score per group, as demonstrated in Figure~\ref{fig:3}.

In order to make the comparison fair, length refers to the number of tokens split in word level. As above mentioned, hybrid word-character level model is one of suitable granularity choices for Chinese-to-English translation. We can find when the length of sentences is below 20, the translation result of this model outperforms the other models to a great extent. But with the length going up, the advantage over other models is diminishing. The character level granularity performs bad for the sentences whose length are below 20. We think the reason may be that when the sentences are short, the representation of sentence in character level cannot express the sentence meaning well. As for BPE method, we find a strange phenomenon. When the number of words in source sentence is from 60 to 80, the translation performance of BPE method is not so good. However, this method can achieve almost 3 BLEU points higher than next-best approach when the source sentence is longer than 80 words. As shown in Figure~\ref{fig:3}, we can see WPM method does not perform well lower than 60 words in source language. But when the length of sentences is between 60 and 80, this method even outperforms the BPE method by up to 5.51 BLEU points. In this experiment, we conclude that subword model is more effective than other models in handling long sentences.

\subsubsection{15k Vocabulary Size}
We concern what the translation results of different translation granularities are on smaller vocabulary size. We also carry out the experiment on Chinese-to-English task of 15k vocabulary size.

\begin{table}
\centering
\caption{Translation results (BLEU score) of 15k vocabulary for Chinese-to-English translation.} \label{c2e15}
\begin{tabular}{p{4.3cm}|p{1.4cm}<{\centering}p{1.4cm}<{\centering}p{1.4cm}<{\centering}p{1.4cm}<{\centering}|p{1.4cm}<{\centering}}
  \hline
  Segmentation (15k)              &    MT03(dev) &  MT04   &  MT05   &   MT06   & Ave \\
  \hline
  \hline
  Word level               &    39.03   &   42.42   &   38.84  & 39.58  & 39.97 \\
  Character level                &    42.60   &   43.60   &   40.85  & 41.29  & 42.09 \\
  Hybrid word-characters level   & 43.58   &      44.25   &   42.29   &   42.37 &  43.12  \\
  BPE method   &    {\bf 44.17}   &    44.89   &    42.79  & {\bf 42.72} &  43.64 \\
  WPM method (raw)  &43.31 & 43.62   & 41.63   & 41.23  & 42.46   \\
  WPM method    & 44.03 &   {\bf 45.15} &  {\bf 43.05} &  42.63  & {\bf 43.72} \\
  \hline
\end{tabular}
\end{table}
Compared to 30k vocabulary size, the translation performance of word level (Row 1) on 15k vocabulary size is reduced by 2.14 BLEU points. However, character level (Row 2) and hybrid word-character level (Row 3) achieve 42.09 and 43.12 BLEU points respectively, which is on par with quality of translation on 30k vocabulary. Both these two models exceed word level to a great extent. We infer the reason is that both character level and hybrid word-character level can represent source side and target side sentences better than the word level even if the vocabulary size is small. For subword model, translation performance of these methods remain almost the same as 30k vocabulary, which is beyond our imagination. We can find in Table~\ref{c2e15}, WPM method (Row 6) outperforms other models, and to our surprise, translation results of both WPM method and WPM methods with raw corpus (Row 5) obtain a higher BLEU points than 30k vocabulary size. We analyze the reason of this phenomenon is that the subword model is not constrained by the vocabulary size. Although the WPM method achieves the best results for the 15k vocabulary size, this method also belongs to subword level translation granularity. We can conclude that subword translation granularity is more suitable for Chinese-to-English translation task.

\subsubsection{60k Vocabulary Size}
In order to make a comparison of these translation granularities on larger vocabulary size, we perform the our experiment of 60k vocabulary size on Chinese-to-English translation task.  

\begin{table}
\centering
\caption{Translation results (BLEU score) of 60k vocabulary for Chinese-to-English translation.} \label{c2e60}
\begin{tabular}{p{4.3cm}|p{1.4cm}<{\centering}p{1.4cm}<{\centering}p{1.4cm}<{\centering}p{1.4cm}<{\centering}|p{1.4cm}<{\centering}}
  \hline
  Segmentation (60k)              &    MT03(dev) &  MT04   &  MT05   &   MT06   & Ave \\
  \hline
  \hline
  Word level               &    42.92   &   44.42   &   41.99  & 42.48  & 42.95 \\
  Character level                &    43.01   &   44.38   &   41.35  & 42.44  & 42.80 \\
  Hybrid word-characters level   & {\bf 43.84}   &  {\bf 45.11}  &  {\bf 43.24}   &   {\bf 43.68} &  {\bf 43.97}  \\
  BPE method   &     43.17   &    44.75   &    42.85  &  43.23 &  43.50 \\
  WPM method (raw)  &40.85 & 42.64   & 38.87   & 40.48  & 40.71   \\
  WPM method    & 43.75 &    44.88 &  41.49 &  41.55  & 42.92 \\
  \hline
\end{tabular}
\end{table}

We can find in Table~\ref{c2e60}, the word and character level (Row 1 and Row 2) on 60k vocabulary size are increased by 1.15 and 1.11 BLEU points respectively compared to 30 vocabulary size. However, to our surprise, all the translation results of subword level granularities on 60k vocabulary are below to the 30k vocabulary size.  With the increase of vocabulary size, we add more fine-grained subword segmentation units into vocabulary. We infer that large amount of subword units do not have beneficial effect on the translation results. As for hybrid word-character level, this method achieves 43.97 BLEU points, which is highest among all the translation granularities on 60k vocabulary size. Compared with Table~\ref{c2e},  hybrid word-character level outperforms the best translation result on 30k vocabulary size (BPE method) by 0.22 BLEU points.

\subsubsection{Different Granularities on Both Sides}
We also conduct experiments that we use different translation granularities on source and target side. In order to carry out the experiments easily, we only compare several granularities pairs.
\begin{table}
\centering
\caption{Translation results (BLEU score) of 30k vocabulary for different granularities on  Chinese-to-English translation.} \label{c2ediff}
\begin{tabular}{p{3.5cm}|p{1.4cm}<{\centering}p{1.4cm}<{\centering}p{1.4cm}<{\centering}p{1.4cm}<{\centering}|p{1.4cm}<{\centering}}
  \hline
  Segmentation (30k)              &    MT03(dev) &  MT04   &  MT05   &   MT06   & Ave \\
  \hline
  \hline
  Word\_BPE               &    41.11   &   43.63   &   40.57  & 41.87  & 41.80 \\
  Word\_Hybrid                &    41.36   &   43.28   &   40.83  & 41.29  & 41.69 \\
  Hybrid\_Word   & 43.53   &      44.77   &   42.69   &   42.56 &  43.39  \\
  Hybrid\_BPE    &    {\bf 44.46}   &    {\bf 45.21}   &    {\bf 43.79}  & {\bf 43.56} & {\bf 44.26} \\
  BPE\_Word  &43.23 & 45.00   & 42.13   & 42.75  & 43.28   \\
  BPE\_Hybrid      & 44.28 &    44.89 &   43.29 &  42.45  &  43.73 \\
  \hline
\end{tabular}
\end{table}

In Table~\ref{c2ediff}, we can find that when the source translation granularity is word level (Row 2 and Row 3), the translation performances are relative poor, even worse than the word level of both sides in Table~\ref{c2e}. As for BPE method on source side, the hybrid word-character on target side obtains 43.73 BLEU points (Row 6), which is close to the best translation result in Table~\ref{c2e}. Hybrid\_BPE method achieves up to 44.26 BLEU points (Row 4), which is even higher than BPE method by up to 0.51 BLEU points. This method can acquire best translation result for Chinese-to-English translation task.

\subsection{Results on English-to-Chinese Translation}
\subsubsection{30k Vocabulary Size} We evaluate different translation granularities on the English-to-Chinese translation tasks, whose results are presented in Table~\ref{e2c}.
\begin{table}
\centering
\caption{Translation results (BLEU score) for English-to-Chinese translation.} 
\label{e2c}
\begin{tabular}{p{4.3cm}|p{1.5cm}<{\centering}p{1.0cm}<{\centering}p{1.0cm}<{\centering}p{1.0cm}<{\centering}p{1.0cm}<{\centering}|p{1.0cm}<{\centering}}
  \hline
  Segmentation (30k)              &   MT03(dev)    & MT04    &  MT05   &  MT06   &   MT08   & Ave\\
  \hline
  \hline
  Word level               & 17.44 & 21.67 & 18.53 & 19.27  & 22.80 &  19.94\\
  Character level                & 18.18   & 20.11 & 17.36 & 18.80 & 23.75 & 19.64  \\
  Hybrid word-characters level   &19.81 & 23.28 & 20.99 & {\bf 21.59} &  {\bf 26.13} & {\bf 22.36}   \\
  BPE method    & 19.43 & 23.23 & 19.77 & 20.24 & 24.30 & 21.39  \\
  WPM method (raw)      & 18.66 & 21.19 & 18.34 & 18.43 & 19.06 & 19.14 \\
  WPM method    & {\bf 20.78} & {\bf 24.05} & {\bf 21.07} & 21.54 & 23.27 & 22.14  \\
  \hline
\end{tabular}
\end{table}

We find that hybrid word-character level (Row 3) granularity obtains significant accuracy improvements over word level and this granularity is also superior to other granularities on large-scale English-to-Chinese translation. BPE method (Row 4) in this task does not perform well as Chinese-to-English task, the translation quality of it is lower than hybrid word-character model by up to 0.97 BLEU points. However, another subword level translation granularity WPM method (Row 6) achieves 22.14 BLEU points, which is near the hybrid word-character level. Although the vocabulary of character level on Chinese side is only 7.2k, it can also obtain 19.64 BLEU points (Row 2), which is on par with translation performance of word level. 

\subsubsection{Different Granularities on Both Sides} As Chinese-to-English translation task, we carry out experiments on English-to-Chinese translation for different granularities. According to Table~\ref{c2ediff}, Hybrid\_BPE and BPE\_Hybrid methods acquire relative higher translation quality than other methods. Therefore, in this section we only use these two methods to test which is most suitable for English-to-Chinese translation task.
\begin{table}
\centering
\caption{Translation results (BLEU score) of 30k vocabulary for different granularities on  English-to-Chinese translation.} \label{e2cdiff}
\begin{tabular}{p{3.2cm}|p{1.5cm}<{\centering}p{1.0cm}<{\centering}p{1.0cm}<{\centering}p{1.0cm}<{\centering}p{1.0cm}<{\centering}|p{1.0cm}<{\centering}}
  \hline
  Segmentation (30k)              &   MT03(dev)    & MT04    &  MT05   &  MT06   &   MT08   & Ave\\
  \hline
  Hybrid\_BPE    &    20.31   &    22.16   & {\bf 20.65}  & 19.87 & 25.65  & 21.73 \\
  BPE\_Hybrid      & {\bf 20.36} &    {\bf 23.35} &   20.03 &  {\bf 20.52}  &  {\bf 26.35} & {\bf 22.12} \\
  \hline
\end{tabular}
\end{table}

Table~\ref{e2cdiff} shows that translation performances of both two methods are below to the Hybrid word-character granularity in Table~\ref{e2c}. BPE\_Hybrid method (Row 2) achieves 22.12 BLEU points, which is higher than Hybrid\_BPE method by 0.39 BLEU points and is near the translation quality of WPM method in Table~\ref{e2c}.

\section{Related Work}
The recently proposed neural machine translation has drawn more and more attention. Most of existing work in neural machine translation focus on handling rare words \cite{Li:2016,Sennrich:2016A,Luong:2015B}, integrating SMT strategies \cite{He:2016,Zhou:2017,wang2017neural,Shen:2016}, designing the better framework \cite{Tu:2016,Luong:2015A,Meng:2016} and addressing the low resource scenario \cite{chengjoint,Zhang:2016A,Sennrich:2016B}. 

As for strategies for dealing with rare and unknown words, a number of authors have endeavored to explore methods for addressing them. Luong et al.~\cite{Luong:2015A} and Li et al.~\cite{Li:2016} propose simple alignment-based technique that can replace out-of-vocabulary words with similar words. Jean et al.~\cite{Jean2014On} use a large vocabulary with a method based on importance sampling.

In addition, another direction to achieve rare words problem in NMT is changing the granularity of segmentation. Chung et al.~\cite{Chung2016A} focus on handling translation at the level of characters without any word segmentation only on target side. Luong et al.~\cite{Luong2016Achieving} propose a novel hybrid architecture that combines the strength of both word and character-based models. Sennrich et al.~\cite{Sennrich:2016A} use BPE method to encode rare and unknown words as sequences of subword units. Wu et al.~\cite{Wu:2016} use both WPM method and hybrid word-character model in their online translation system. However, there is no study that shows which translation granularity is suitable for translation tasks involving Chinese language.
Our goal in this work is to make an empirical comparison of different translation granularities for bidirectional Chinese-English translation tasks.

\section{Conclusion}

In this work, we provide an extensive comparison for translation granularities in Chinese-English NMT, such as word, character, subword and hybrid word-character. We have also discussed the advantages and disadvantages of various translation granularities in detail. For the same granularity on both sides, the experiments demonstrate that the subword model best fits Chinese-to-English translation with the vocabulary that is not so big, while the hybrid word-character approach obtains the highest performance on English-to-Chinese translation. In addition, experiments on different granularities show that Hybrid\_BPE method can acquire best result for Chinese-to-English translation task.

\section*{Acknowledgments}
%
The research work has been funded by the Natural Science Foundation of China under Grant No. 61333018 and No. 61402478, and it is also supported by the Strategic Priority Research Program of the CAS under Grant No. XDB02070007.

\bibliography{cwmt}
\bibliographystyle{splncs03}

\end{CJK*}
\end{document}